\documentclass[10pt,twocolumn,letterpaper]{article}

\usepackage[pagenumbers]{cvpr} 

\usepackage{graphicx}
\usepackage{amsmath}
\usepackage{amssymb}
\usepackage{booktabs}

\usepackage[pagebackref,breaklinks,colorlinks]{hyperref}

\usepackage[capitalize]{cleveref}
\crefname{section}{Sec.}{Secs.}
\Crefname{section}{Section}{Sections}
\Crefname{table}{Table}{Tables}
\crefname{table}{Tab.}{Tabs.}

\def\cvprPaperID{7596} 
\def\confName{ICCV}
\def\confYear{2025}

\begin{document}

\title{FocusChat: Text-guided Long Video Understanding via Spatiotemporal Information Filtering}

\author{{Zheng Cheng, Rendong Wang, Zhicheng Wang}\\
YITION.AI\\
\textit{\{zheng.cheng, rendong.wang, zhicheng.wang\}@yition.ai}
}

\maketitle

\begin{abstract}
Recently, multi-modal large language models have made significant progress. However, visual information lacking of guidance from the user's intention may lead to redundant computation and involve unnecessary visual noise, especially in long, untrimmed videos. To address this issue, we propose FocusChat, a text-guided multi-modal large language model~(LLM) that emphasizes visual information correlated to the user's prompt. In detail, Our model first undergoes the semantic extraction module, which comprises a visual semantic branch and a text semantic branch to extract image and text semantics, respectively. The two branches are combined using the Spatial-Temporal Filtering Module~(STFM). STFM enables explicit spatial-level information filtering and implicit temporal-level feature filtering, ensuring that the visual tokens are closely aligned with the user's query. It lowers the essential number of visual tokens inputted into the LLM. FocusChat significantly outperforms Video-LLaMA in zero-shot experiments, using an order of magnitude less training data with only 16 visual tokens occupied. It achieves results comparable to the state-of-the-art in few-shot experiments, with only 0.72M pre-training data.
\end{abstract}

\section{Introduction}
\begin{figure}
  \centering
   \includegraphics[width=\linewidth]{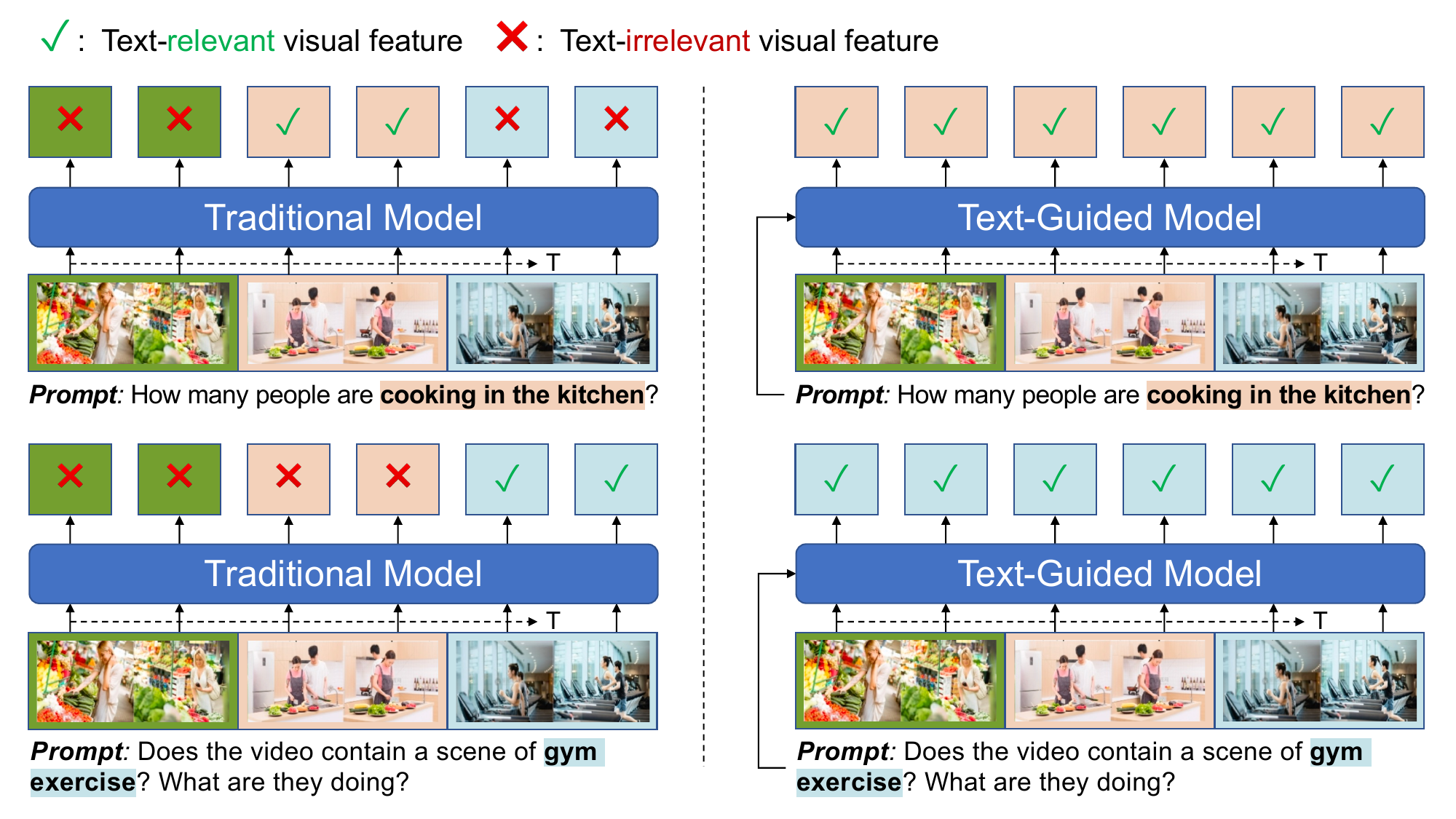}

   \caption{This illustration contrasts traditional and text-guided models. \textbf{Left:} The traditional model interprets visual patches directly into tokens for LLMs without considering specific frames or areas of interest. As a result, whether the inquiry belongs to a ``kitchen" or a ``gym," the model consistently produces the same tokens and applies uniform attention to all details in the scene, potentially increasing the cognitive burden on the LLMs. \textbf{Right:} The text-guided model utilizes prompts to identify the most relevant visual cues and generates adaptive tokens, thereby improving the LLMs' capacity to comprehend and interpret visual information.}
   \label{fig:fig0}
\end{figure}
Large Language Models (LLMs) \cite{chiang2023vicuna,chung2024scaling,touvron2023llama} have emerged as powerful tools in the realm of natural language processing, achieving substantial success through extensive pre-training on vast amounts of textual data. Notable examples include GPT \cite{chatgpt} and LLaMA\cite{touvron2023llama,touvron2023llama2}, which excel in generative and discriminative tasks within a cohesive framework. Recently, there has been an increasing trend in applying LLMs to multimodal tasks, highlighting their potential in areas such as image captioning\cite{mokady2021clipcap,rotstein2024fusecap,liu2024image} and question-answering\cite{jin2024chatunivi,li2023videochat,maaz2023videochatgpt,song2024moviechat,zhang2023videollama,cheng2024videollama2} that utilize various visual inputs. However, understanding long videos poses unique challenges\cite{xu2024slowfast,wu2024longvideobench}, especially in answering questions about content that spans several minutes. 
This task is highly valuable because long-form videos, ranging from educational tutorials to feature films, play a crucial role in our everyday life. 
\begin{figure*}
  \centering
   \includegraphics[width=\linewidth]{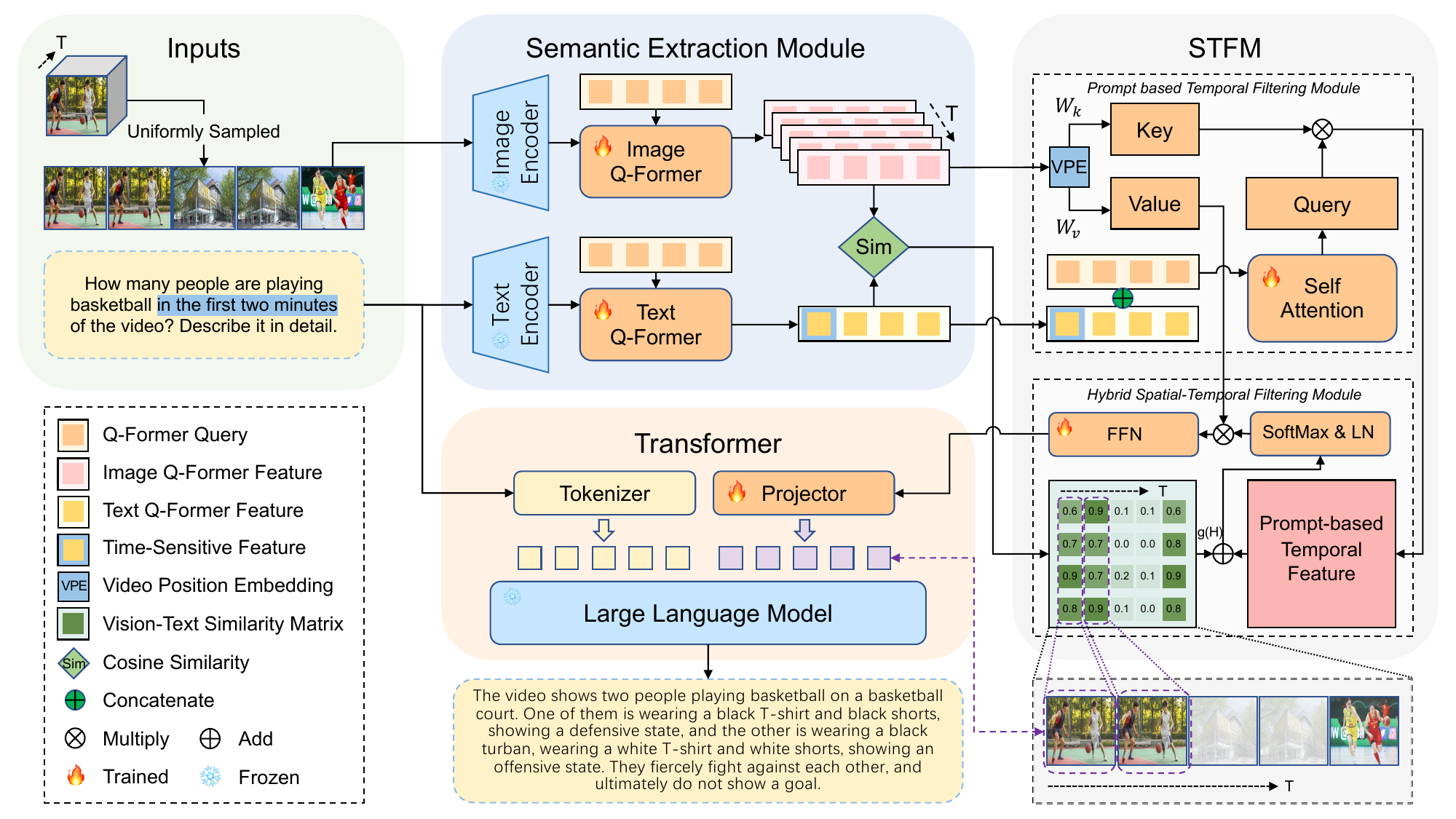}

   \caption{The overall architecture of FocusChat: uniformly sampled frames are input into the vision semantic branch, which consists of an image encoder and an image Q-Former. Simultaneously, the user's query is input into the text semantic branch to extract rich semantic representations. Finally, these are fused in STFM to achieve both spatial-level and temporal-level filtering of visual information. The output of STFM is projected as visual tokens, which are fed into the LLM along with the text tokens.}
   \label{fig:fig1}
\end{figure*}
When people want to seek specific knowledge in a long video, a quick way is to first identify and locate relevant segments of the video according to the intention. Then, we focus on the content of those segments, paying attention to highly related information rather than treating every frame in the same manner. Whereas most existing video understanding models\cite{cheng2024videollama2,song2024moviechat,maaz2023videochatgpt,ren2024timechat,tan2024koala,song2024moviellm} treat all input frames indiscriminatingly, leading to output visual tokens that contain significant redundant information, especially in untrimmed videos containing diverse scenes. These models either scale up or increase the number of visual tokens provided to the LLM. Unfortunately, integrating long visual sequences into Large Language Models (LLMs) raises additional complexities. As shown in \cref{fig:fig0} , no matter whether the query is ``How many people are cooking in the kitchen?" or ``Does the video contain a scene of gym exercise? What are they doing?" the traditional model transforms the video into identical tokens, resulting in redundancy of information. In contrast, the text-guided model extracts visual information based on the query content, enhancing the model's capabilities via concentration.

In order to address the aforementioned issues, we propose a text-guided model for long video understanding. The core idea is to extract visual content that closely aligns with the user's prompt. We leverage the image Q-Former from Video-LLaMA \cite{zhang2023videollama}  to extract visual and text information. The visual and textual semantic representations are then fed into a Spatial-Temporal Filtering Module~(STFM). As shown in~\cref{fig:fig1}, the two submodules of STFM, which are the prompt-based temporal filtering~(PBTF) module and the hybrid spatial-temporal filtering~(HSTF) module, together make the generated visual tokens closely related to the semantics of the user's query.

The PBTF module extracts visual features consistent with the semantics of the prompt, and the HSTF module adopts a vision-text similarity matrix to filter visual tokens from both spatial and temporal perspectives. As shown in \cref{fig:fig0} , when the user asks: ``How many people are playing basketball in the first two minutes?", the model's response has nothing to do with scenes after the first two minutes and scenes not containing playing basketball. PBTF adopts time-sensitive tokens of the words ``first two minutes" as queries to filter the keys and values of visual information. Meanwhile, HSTF omits scenes and regions with low correlations with the query via the vision-text similarity matrix. Ultimately, only features corresponding to the query are used as visual tokens. A novel vision position embedding approach is proposed to facilitate the filtering process.

\noindent\textbf{We summarize our contributions as follows:}
\begin{itemize}
\item We propose a novel approach called FocusChat with a Spatial-Temporal Filtering Module to align the visual tokens properly with the prompt. Specifically, two sub-modules of STPM are proposed~(Prompt-based temporal filtering module and hybrid spatial-temporal filtering module) to generate efficient and effective visual tokens.
\item FocusChat achieves competitive results in zero-shot and few-shot experiments with only an order of magnitude fewer data and even 16 visual tokens occupied, making it much easier to use in practice.
\item We conducted thorough ablation experiments on each module in STFM. The results demonstrate that the proposed method is both efficient and effective.

\end{itemize}

\begin{figure*}
  \centering
   \includegraphics[trim=0 0 0 60, clip,width=0.9\linewidth]{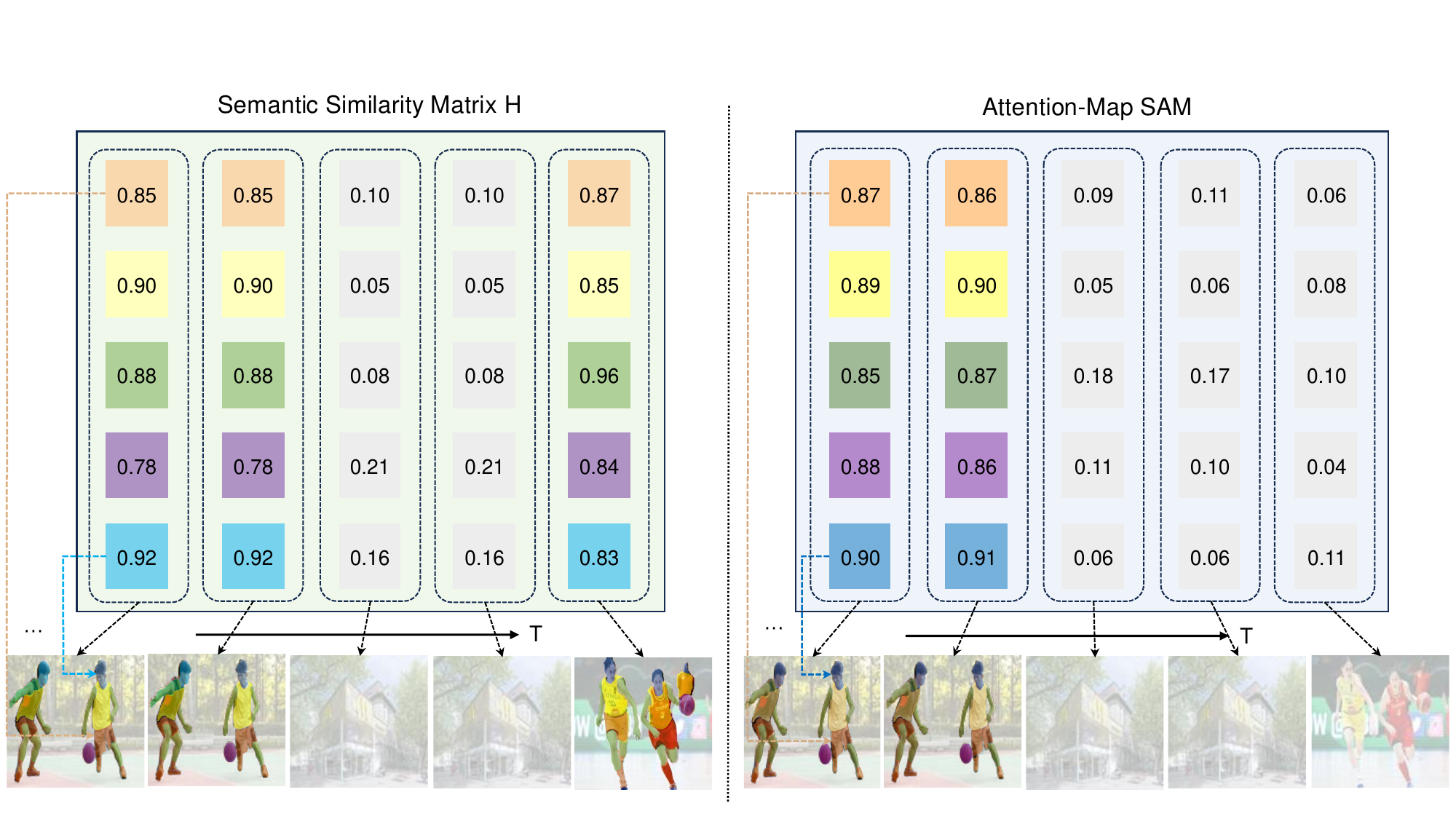}

   \caption{When asking a question about a five-minute video, such as ``How many people are playing basketball in the first two minutes?", the semantic similarity matrix H and the attention-map SAM diagram in STFM are presented as indicative illustrations.}
   \label{fig:fig_add}
\end{figure*}

\section{Related Work}
\label{sec:related_work}
\textbf{Large Language Models }(LLMs). Language is a key skill for expression and communication in humans, which begins to develop in early childhood and continues to evolve throughout life \cite{pinker2003language,hauser2002faculty}. Enabling machines to read, write, and communicate like humans has long been a challenging research goal\cite{turing1950computing} that is significant for human development. LLM is currently a popular way to implement such functionality. Typically, large language models (LLMs) refer to transformer language models \cite{vaswani2017attention} that contain hundreds of billions (or more) of parameters, which are trained on massive text data\cite{shanahan2024talking}, such as GPT-3 \cite{mann2020language}, PaLM \cite{chowdhery2022palm}, Galactica\cite{taylor2022galactica}, and LLaMA \cite{touvron2023llama}. LLMs exhibit strong capacities to understand natural language and solve complex tasks (via text generation).

\textbf{Vision Large Language Models }(vLLMs). The integration of the perceptual abilities of the vision models \cite{kirillov2023segment,mokady2021clipcap,oquab2023dinov2,radford2021learning} with the reasoning capabilities of LLMs has given rise to Vision Large Language Models (vLLMs) \cite{yang2024doraemongpt,hu2024minicpm,li2024densefusion}. VLLMs encompass both Image-Language and Video-Language models. This approach transforms visual signals into tokens interpretable by LLMs. Image-Language models integrate powerful pre-trained language models with image encoders to enhance multi-modal reasoning capabilities\cite{li2023blip,liu2024visual,zhu2023minigpt}. For instance, Flamingo \cite{alayrac2022flamingo} connects state-of-the-art vision-only and language-only models to excel in few-shot learning tasks. BLIP-2\cite{li2023blip} introduces a lightweight querying transformer to bridge the gap between frozen image encoders and language models, achieving strong performance with fewer trainable parameters. LLaVA \cite{liu2024visual} utilizes a simple linear layer to project image features into the text embedding space, effectively fine-tuning language models for improved outcomes. MiniGPT-4\cite{zhu2023minigpt} builds on BLIP-2\cite{li2023blip}by gathering a large dataset of image-text pairs, enhancing language generation. Video-language models have evolved from image-language models like Flamingo \cite{alayrac2022flamingo} and BLIP-2\cite{li2023blip}, which flatten spatio-temporal features into 1D sequences but struggle to capture temporal dynamics. Models such as Video-LLaMA\cite{zhang2023videollama}  add video querying transformers to enhance temporal understanding, while Video-ChatGPT\cite{maaz2023video} averages spatial-level features for video representation. ChatVideo\cite{wang2023chatvideo} uses tracklets annotated with textual descriptions and employs ChatGPT for various tasks, while VideoChat \cite{li2023videochat} generates action and object annotations for LLM reasoning.

\textbf{Long Video Understanding}. Long video understanding poses significant challenges in computer vision, primarily because it requires capturing long-range patterns in videos that often exceed 30 seconds. A typical strategy\cite{balavzevic2024memory,wu2022memvit,cheng2022xmem} involves maintaining a memory bank to store historical information and utilizing parametric\cite{wu2022memvit} or non-parametric\cite{balavzevic2024memory} compression modules to manage this efficiently. Recent methodologies \cite{kahatapitiya2024language,he2024ma,ranasinghe2024understanding} have incorporated language as an intermediary for enhanced comprehension, segmenting extended videos into shorter clips, producing textual descriptions for each segment, and subsequently utilizing large language models (LLMs) to consolidate these captions for further analysis. However, these methods are cumbersome and lack conciseness. Some have adopted streaming methods\cite{he2024ma,qian2024streaming,zhou2024streaming,ren2024timechat} to process long videos. Although these methods appear promising, indiscriminately handling all video frames inevitably leads to information redundancy. \cite{tan2024koala} is a key-frame-based model, distinct from previous methods. Although it appears to treat each frame differently, the key frame is fixed and unrelated to the prompt. Our proposed text-guided approach effectively addresses this issue. There are very few video understanding models based on this method, with \cite{qian2024streaming} being one of them. However, it has several drawbacks, including a complex model structure, time-consuming computations, and an inability to filter information at the frame level or even at the region level. In contrast, our FocusChat effectively overcomes these limitations.

\section{Method}
\label{sec:method}
We present FocusChat, which consists of the semantic extraction module and STFM. The former includes the vision branch and the text branch, which will be introduced in \cref{sec:3.1}. These branches are subsequently merged in STFM, which will be discussed in \cref{sec:3.2}.

\subsection{Semantic Extraction Module}
\label{sec:3.1}
The semantic extraction module includes vision and text semantic branches. It extracts user prompt and visual features. These semantic features are then fed into STFM, which filters the image semantics using the text semantics to get visual features closely related to intention.
\subsubsection{Visual Semantic Branch}
The visual semantic branch extracts the semantic information from each image. It consists of a frozen pre-trained image encoder and a trainable image Q-Former. A video consists of $T$ frames is represented as $\mathbf{V}\in \mathbb{R}^{T\times H\times W\times 3}$.  The input frames are passed into a pre-trained visual encoder, i.e., ViT, to obtain the visual frame features $\mathbf{V_f}\in \mathbb{R}^{T\times N_v\times C}$, where $N_v$ represents the number of patches, and $C$ denotes the number of feature channels. Subsequently, an image Q-Former further compresses the frame features. As \cref{fig:fig1} illustrates, the image Q-Former takes as input \textit{M} learnable queries of length \textit{D}. These queries interact with the frame features via cross-attention and update the initial queries to output the final \textit{M} visual semantic vectors of length \textit{D}, denoted as $\mathbf{V_q}\in \mathbb{R}^{T\times M\times D}$. Each visual vector contains semantic information at the region level or frame level. For example, the visual semantics extracted from the video in \cref{fig:fig0} may include elements such as men, women, vegetables, treadmills, etc.
\subsubsection{Textual Semantic Branch}
The text semantic branch is designed to semantically encode the user’s input prompt $P$. A pre-trained CLIP\cite{radford2021learning} model encodes the user prompt, resulting in prompt features $\mathbf{P_f}\in \mathbb{R}^{N_p\times D}$ , where $D$ denotes the dimension of text embedding, and $N_p$ is the text token number.  We then employ a Q-Former, similar to the visual semantic branch, to encode $\mathbf{P_f}$. In this process, $\mathbf{P_f}$ is projected as keys and values. By utilizing \textit{M} trainable query embeddings, we obtain \textit{M} text vectors, denoted as $\mathbf{P_q}\in \mathbb{R}^{M\times D}$. Each textual vector contains semantic information at the word or subword level.  For example, the prompt ``How many people are cooking in the kitchen?" in \cref{fig:fig0} may be parsed into semantic vectors representing person, cooking, kitchen, etc.


\subsection{Spatial-Temporal Filtering Module~(STFM)}
\label{sec:3.2}
\subsubsection{Prompt Based Temporal Filtering Module}
 As mentioned earlier, previous work lacks information filtering at the frame level. Therefore, we first propose a “Prompt-based Temporal Filtering Module” to assist our model in filtering information at the temporal level. In this module, we first add a temporal position encoding to $\mathbf{V_q}$. Unlike the trainable position encoding used in the original Video-LLaMA, we design a position encoding based on the transformer's\cite{vaswani2017attention} model. Experiments indicate that the revised position encoding is more effective. The traditional transformer's\cite{vaswani2017attention} position embedding is shown in \cref{eq:dd}, where \textit{pos} represents the position, specifically the frame index here. \textit{i} denotes the dimension, and \textit{d} is the feature dimension. We improve upon it by substituting \textbf{VPE} for \textit{pos}, where $VPE=pos\cdot(S/T)$, and $S$ is a constant, T is the number of input frames. The ablation experiments demonstrate that our VPE enhances the effectiveness of STFM.
\begin{equation}
\small
\begin{split}
 PE_{(pos,2i)} = sin(pos / 10000^{2i/d}) \\
PE_{(pos,2i+1)} = cos(pos / 10000^{2i/d})
\end{split}
\label{eq:dd}
\end{equation}

The $\mathbf{V_q}$ added with temporal position encoding are then projected into keys and values using $\mathbf{W_k}$ and $\mathbf{W_v}$, respectively. Meanwhile, another set of N trainable queries is concatenated with the text semantic features $\mathbf{P_q}$ and used as input to a self-attention module to incorporate time-sensitive information into the queries. This helps our model extract visual features that aligned with the temporal information during hybrid spatial-temporal filtering cross-attention, enhancing the signal-to-noise ratio. For example, the visual features of the basketball scene in the last minute of \cref{fig:fig1} will also be filtered out as redundant information. Concretely , we only use the output of the self-attention corresponding to the trainable queries and linearly project it as the Query in \cref{fig:fig1}, which is referred to as $\mathbf{V_s}\in \mathbb{R}^{N\times D}$ in the following text. Subsequently, $\mathbf{V_s}$ is multiplied by the keys, which contain temporal position encoding information, resulting in the prompt-based temporal feature. This prompt-based temporal feature, and the values, which also contain temporal position encoding information, will be fed into the next module to provide features that emphasize temporal filtering.

\subsubsection{Hybrid Spatial-Temporal Filtering Module}
In addition to the features that emphasize temporal filtering, we also need to obtain visual features that are semantically close to the user’s query, we first compute the semantic similarity matrix by calculating the similarity between text and visual semantic vectors. The similarity matrix is then used to guide the hybrid spatial-temporal filtering (HSTF) module for visual information filtering. The filtering process is explicitly applied to the video in both spatial and temporal dimensions. Spatial filtering is performed row by row, as shown in \cref{fig:fig_add}, while temporal filtering is applied column by column.

\textbf{Semantic Similarity Matrix}. We compute the cosine similarity between the text semantic feature $\mathbf{P_q}$ and the visual semantic feature $\mathbf{V_q}$ to get the similarity matrix $\mathbf{H}\in\mathbb{R}^{T \times M }$. As shown in \cref{eq:aa}, \textit{sim} indicates cosine similarity and normalization, \textit{t} is the frame index, and \textit{i} is the semantic feature index. Since text semantics can achieve word-level granularity and visual semantics can achieve region-level granularity, therefore $\mathbf{H}$ could represents fine-grained similarity. For example, in \cref{fig:fig_add}, each element of the semantic similarity matrix H represents a specific region in the image, such as an arm, body, or basketball. It serves as a guide for filtering the visual spatial-temporal features.

\begin{equation}
\small
H_{t,i}=sim(V_{q_{t,i}},P_{q_{i}})=(\frac{V_{q_{t,i}}\cdot P_{q_{i}}}{\parallel V_{q_{t,i}}\parallel \times \parallel P_{q_{i}}\parallel}+1)\times\frac{1}{2}
\label{eq:aa}
\end{equation}

The Video Q-Former in Video-LLaMA \cite{zhang2023videollama} treats all frames indiscriminately. We improved the Video Q-Former by enabling it to receive a semantic similarity matrix to filter visual features based on its values. This process occurs in cross-attention of STFM, allowing FocusChat to extract accurate vision semantics related to the user's question more effectively and explicitly. As a result, it enhances the accuracy and generalization ability of FocusChat while reducing the load on the LLM.

\textbf{Hybrid Spatial-Temporal Filtering Cross-Attention}. We aim to produce an attention map that is closely related to the semantic similarity matrix. Our hybrid spatial-temporal  filtering module based attention-map $\mathbf{SAM}\in\mathbb{R}^{N\times TM}$ is given in \cref{eq:bb}, which is equivalent to $SAM=softmax(\alpha log(H)+QK^T/\sqrt{d})$, where $\alpha$ is a constant greater than or equal to 0, Q is the Query of PBTF, and K is the Key of PBTF. So $g(H)=\alpha log(H)$, as shown in \cref{fig:fig1}. The output of the hybrid spatial-temporal filtering  module is denoted as $\mathbf{Z}\in\mathbb{R}^{N\times d}$.  \cref{eq:bb}  and  \cref{eq:cc}  complete most of the spatiotemporal information filtering. Therefore, the redundant information filtered in  $\mathbf{V_q}$ alleviates the pressure on the number of vision tokens sent to LLM.
\begin{equation}
\small
SAM_{j,i}=\frac{H_i^\alpha e^{V_{s_i}W_q(V_{q_i}W_k)^T/\sqrt{d}}}{\sum_{i=1}^{TM}H_i^\alpha e^{V_{s_i}Wq(V_{q_i}W_k)^T/\sqrt{d}}}
\label{eq:bb}
\end{equation}

\begin{equation}
\small
    Z_{j}=LayerNorm(\sum_{i=1}^{TM}H_i^\beta SAM_{j,i}V_{q_i}W_v)
    \label{eq:cc}
\end{equation}

Through the HSTF process, we obtain visual-semantic features highly relevant to the user's prompt. As shown in \cref{fig:fig_add}, useful semantic information from all frames is extracted, while irrelevant visual-semantic features are filtered out.

 \begin{table}
  \centering
  \begin{tabular}{@{}lc@{}cll}
    \toprule
    Modality& dataset& Original& Used&Ratio\\
    \midrule
    Video-Text& webvid\cite{bain2021frozen}&  10M& 0.62M&6.2\%\\
  Image-Text&  CC-3M\cite{changpinyo2021conceptual}& 3M& 0.10M&3\%\\
     total& -& 13M& 0.72 M&5.5\%\\
    \bottomrule
  \end{tabular}
\caption{Zero-shot pre-training data details.}
\label{tab:pretrain_data}
\end{table}

\begin{table}
  \centering
  \begin{tabular}{@{}lc@{\hspace{1em}}c}
    \toprule
    Hyper-parameter& first stage& second stage\\
    \midrule
    $\mathbf{\alpha }$ in STFM& \multicolumn{2}{c}{1}\\
 $\mathbf{\beta }$ in STFM& \multicolumn{2}{c}{0}\\
 $S$ in \textbf{VPE}& \multicolumn{2}{c}{500}\\
 Number of video tokens& \multicolumn{2}{c}{32}\\
 Number of all Q-Former queries& \multicolumn{2}{c}{32}\\
 Number of input frames T& \multicolumn{2}{c}{15}\\
 Max text length& \multicolumn{2}{c}{2048}\\
    Epochs& \multicolumn{2}{c}{1}\\
 Batch size& \multicolumn{2}{c}{128}\\
 Weight decay& \multicolumn{2}{c}{0.05}\\
 AdamW $\beta$& \multicolumn{2}{c}{(0.9, 0.999)}\\
 Warm-up learning rate& \multicolumn{2}{c}{1e-6}\\
 LLM& \multicolumn{2}{c}{LLaMA2-7B}\\
    Learning rate& 1e-4&3e-5\\
    \bottomrule
  \end{tabular}
\caption{Hyper-parameters of two training stages.}
\label{tab:example}
\end{table}
\subsection{Model Training}
\subsubsection{Zero-Shot Training}
 To train FocusChat, we design a two-stage paradigm. We use a total of 1.5 million data samples for zero-shot training. \textbf{In the first stage}, The pre-training data is detailed in \cref{tab:pretrain_data}. We utilize part of the picture description pairs from CC-3M\cite{changpinyo2021conceptual} and video description pairs from WebVid\cite{bain2021frozen}, totaling approximately only 0.72 million pairs. Since there are no user instructions at this stage, we construct various instruction templates, such as ``Describe this video in detail," and randomly select templates to generate user input during training. The components that can be optimized in this stage include the image Q-Former, text Q-Former, projector, and STFM. \textbf{In the second stage}, we retain the trainable modules from the previous stage but use different training data VideoChat2-IT\cite{li2024mvbench}, which includes NExTQA\cite{xiao2021next}, TextVR\cite{wu2025large}, CLEVRER\cite{yi2019clevrer}, TGIF\cite{li2016tgif}, Kinetics-710\cite{kay2017kinetics}, EgoQA\cite{fan2019egovqa}, ShareGPT4Video\cite{chen2024sharegpt4video}, etc. We sample a small subset of data from VideoChat2-IT as the fine-tuning dataset, totaling approximately 0.8M, reformulating their instructions to fit the specific structure of FocusChat. 
 \begin{table*}
  \centering
  \begin{tabular}{@{}lcc@{\hspace{1em}}ccc}
    \toprule
    method &PT& Activitynet-QA& MSVD-QA& MSRVTT-QA&Next-QA\\
    \midrule
    JustAsk\cite{yang2021just}&69M&   38.9 & 47.5& 41.8& 50.8\\
 FrozenBiLM\cite{yang2022zero}&400M& 43.2 & 54.8& 47.0& -\\
 Singularity\cite{lei2022revealing}&17M&  44.1& -& 43.5& -\\
 VIOLETv2\cite{fu2023empirical}&5M& 44.5&  54.7& -& -\\
 GiT\cite{wang2022git}&800M& 43.2&  56.8& -& -\\
 mPLUG-2\cite{xu2023mplug}&17M& 48.0& 58.1& -& -\\
 UMT-L\cite{li2023unmasked}&25M& 47.9&  55.2& 47.1& -\\
 VideoCoCa\cite{yan2022videococa}&4.8B& \textbf{56.1}&  56.9& 46.3& -\\
 MA-LMM\cite{he2024ma}&-& \underline{49.8}& \underline{60.6}& \underline{48.5}& -\\
 SeViLA\cite{yu2024self}& 129M& -& -& -& \textbf{73.4}\\
 HiTeA\cite{ye2023hitea}&5M& 45.1& 55.6 & 45.4&  63.1\\
 IGV\cite{li2022invariant}&-& -& 40.8& 38.3&  51.3\\
 HQGA\cite{xiao2022video}&-& -& 41.2& 38.6& 51.8\\
 FocusChat &0.72M& 42.5& 54.6& 45.4& \underline{68.2}\\
    FocusChat* &0.72M& 49.4*& \textbf{63.7*}& \textbf{54.4*}&-\\
    \bottomrule
  \end{tabular}
  \caption{The few-shot evaluation results of various models on ActivityNet-QA, MSVD-QA, MSRVTT-QA, and Next-QA. * indicates GPT-3.5-turbo-0613 evaluation, while no * means non-GPT-3.5-turbo-0613 evaluation. Bold represents the first place, and the underscore indicates the second place. PT refers to the number of pre-training datasets.}
  \label{tab:few-shot}
\end{table*}
\begin{table*}
  \centering
  \begin{tabular}{@{}lc@{\hspace{1em}}ccc}
    \toprule
    benchmark& Activitynet-QA & MSVD-QA& MSRVTT-QA&MovieChat-1K\\
 method& acc/score& acc/score& acc/score&acc/score\\
    \midrule
    Video-LLaMA& 12.4/1.1& 51.6/2.5& 29.6/1.8&51.7/2.6\\
    FocusChat& 33.2/3.1& 52.4/3.4& 46.7/3.2&60.0/3.5\\
    difference value& +20.8/+2.0& +0.8/+0.9& +17.1/+1.4&+8.3/+0.9\\
    \bottomrule
  \end{tabular}
  \caption{The zero-shot evaluation results of FocusChat and Video-LLaMA on the ActivityNet-QA, MSVD-QA, MSRVTT-QA, and MovieChat-1K datasets, with the ``difference value" representing the performance gap where FocusChat exceeds Video-LLaMA.
}
  \label{tab:zero-shot}
\end{table*}
\begin{table}
  \centering
  \begin{tabular}{@{}lc}
    \toprule
    method&Accuracy\\
    \midrule
    Video-LLaMA&45.07\\
    baseline(Video-LLaMA w/ \textbf{VPE}+ln)&47.34\\
 ours(baseline+$\alpha$=0+$\beta$=1)&48.17\\
 ours(baseline+$\alpha$=0.5+$\beta$=1)&48.55\\
 ours(baseline+$\alpha$=1+$\beta$=1)&48.85\\
 ours(baseline+$\alpha$=1+$\beta$=0)&48.93\\
 ours(baseline+$\alpha$=1+$\beta$=1+PBTF)&49.05\\
 \textbf{FocusChat(baseline+$\alpha$=1+$\beta$=0+PBTF)}&\textbf{49.40}\\
    FocusChat wo \textbf{VPE}&49.29\\
    \bottomrule
  \end{tabular}
  \caption{Ablation of the few-shot results for each module parameter of FocusChat on ActivityNet-QA benchmark.}
  \label{tab:ablution}
\end{table}
\begin{table}
  \centering
  \begin{tabular}{@{}lccc}
    \toprule
    method  &PT/FT&NT&Acc/score\\
    \midrule
    Video-LLaMA&Millions/Millions&32&12.4/1.1\\
    FocusChat  &0.72M/0.8M&32&33.2/3.1\\
    FocusChat(16)&0.72M/0.8M&16&27.7/2.8\\
    \bottomrule
  \end{tabular}
  \caption{Ablation of the zero-shot results for the number  of  vision tokens on ActivityNet-QA benchmark. PT represents the amount of pretraining data, FT represents the amount of fine-tuning data, and NT represents the number of vision tokens to LLM.}
  \label{tab:num_token}
\end{table}
\subsubsection{Few-Shot Training}
To more comprehensively evaluate the model's performance, we conducted few-shot training in addition to zero-shot experiments. The few-shot training is based on the zero-shot model. We use the semantic extraction module of the zero-shot fine-tuned model and the remaining modules of the pre-trained model as the initialization for few-shot training. The parameters of the semantic extraction module are fixed, and only the STFM module and the projection layer are optimized to ensure consistency and stability in semantic extraction. In this phase, we train on each benchmark separately, with the data details provided in \cref{sec:3.4.2}.
\section{Experiments}
\subsection{Implementation Details}
We use the  ViT-G/14 from EVA-CLIP\cite{fang2023eva} as the image encoder and CLIP\cite{radford2021learning} as the text encoder to ensure that the extracted text features and corresponding visual features are aligned in the embedding space. The image Q-Former and text Q-Former are initialized with the InstructBLIP's\cite{2023InstructBLIP} checkpoint, while the STFM is randomly initialized. We use the open-source LLaMA2 (7B) model as the LLM. In STFM, the parameters $\mathbf{\alpha }$ and $\mathbf{\beta}$ are set to 1 and 0, respectively, and the number of queries for all Q-Formers is 32. Hyper-parameters for all two zero-shot training stages are provided in \cref{tab:example}. The few-shot training parameters are the same as those for zero-shot, except for the learning rate, which is set to 5e-5. Each few-shot training sample also includes the instruction: ``Please answer as briefly as possible."

\subsection{Datasets }
\label{sec:3.4.2}
Our zero-shot model evaluation benchmarks include ActivityNet-QA \cite{yu2019activitynet}, MSVD-QA\cite{xu2017video}, MSRVTT-QA\cite{xu2017video}, and MovieChat-1K\cite{song2024moviechat}, all of which are open-ended visual question-answering datasets. Except for MovieChat-1K\cite{song2024moviechat} with an average duration of 8 minutes, the video durations of the other datasets are around 1 to 2 minutes. For these evaluations, we use the widely adopted GPT-3.5-turbo-0613. For the few-shot experiments, we conduct training and testing on ActivityNet-QA\cite{yu2019activitynet}, MSVD-QA\cite{xu2017video}, MSRVTT-QA, and Next-QA\cite{xu2017video}. Next-QA\cite{xiao2021next} has an average duration of 42 seconds and is a multiple-choice dataset. Since there is no standardized evaluation method for open-domain visual question-answering datasets in few-shot tasks, we used two evaluation approaches for all datasets except Next-QA\cite{xiao2021next}: one with GPT-3.5-turbo-0613 and another with a strict evaluation method, where a prediction is considered correct only if it exactly matches the ground truth.
\subsection{Main Results }
\textbf{Zero-Shot Result. } For the quantitative experiments, we use a very small training dataset of approximately 1.5 million samples, which makes a direct comparison with models trained on tens of millions of samples somewhat unfair. Our model builds upon and enhances the Video-LLaMA structure to achieve better video understanding. Thus, we mainly compared FocusChat with Video-LLaMA. Video-LLaMA uses a dataset of tens of millions of samples, whereas we use only about one-tenth of that amount. Despite this, FocusChat outperforms Video-LLaMA across all benchmarks.

\begin{figure*}
  \centering
   \includegraphics[trim=0 0 0 70, clip,width=\linewidth]{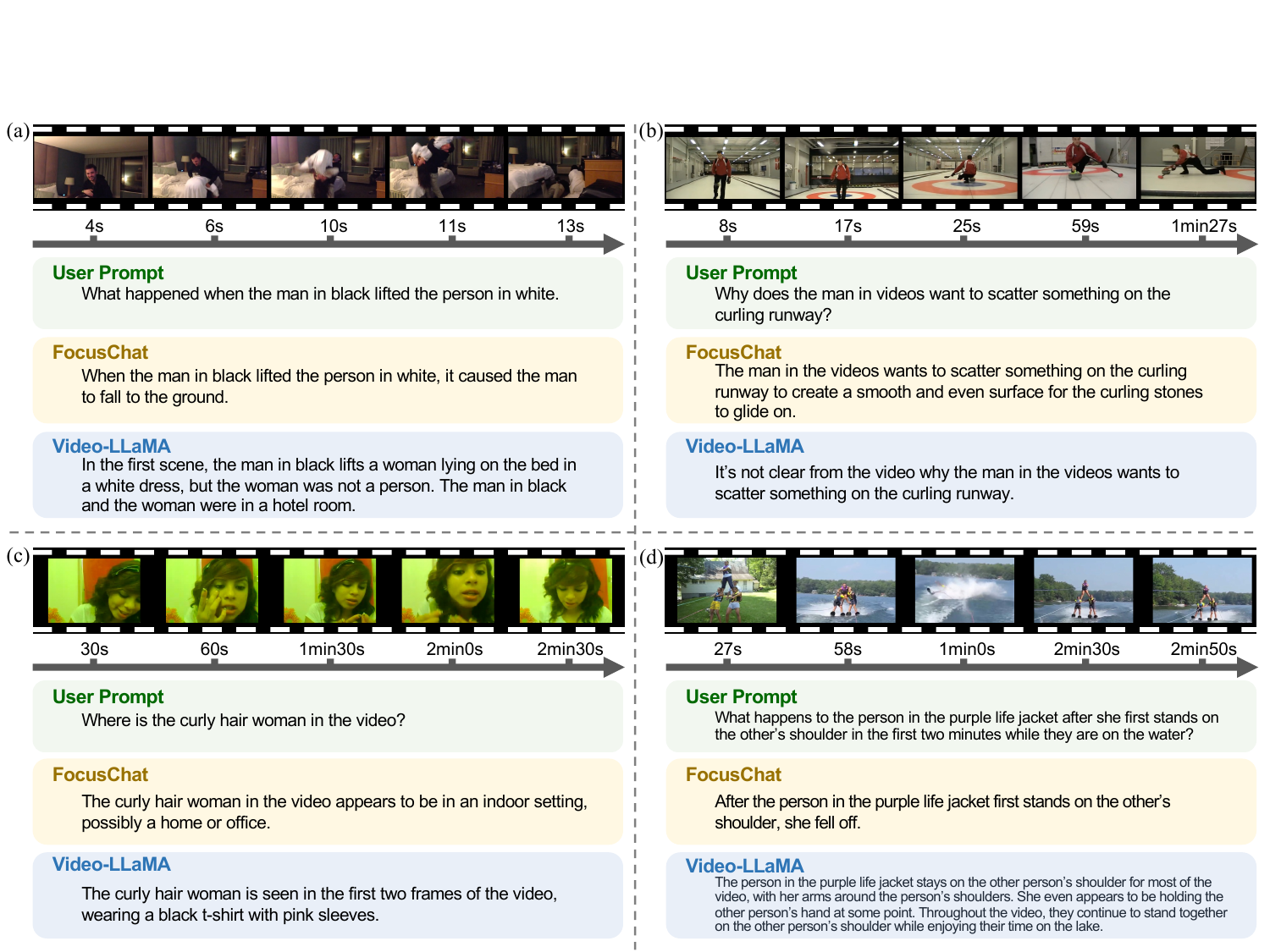}

   \caption{Qualitative result comparison between Video-LLaMA and FocusChat.}
   \label{fig:fig2}
\end{figure*}
As shown in  \cref{tab:zero-shot}, on the ActivityNet-QA dataset, FocusChat’s accuracy exceeds Video-LLaMA by 20.8, with a score improvement of around 2. On the MSRVTT-QA dataset, FocusChat’s accuracy surpasses Video-LLaMA by 17, with a score increase of 1.4. For MovieChat-1K, we only test the global mode, and FocusChat's accuracy outperforms Video-LLaMA by 8.3, demonstrating FocusChat's superior ability to understand long videos.On the MSVD-QA dataset, FocusChat matches Video-LLaMA in accuracy, with its score 0.94 higher.

FocusChat's average score across all datasets is approximately 3.3, reflecting the high quality of its responses. For the qualitative experiments, we compared the zero-shot performance of Video-LLaMA and FocusChat, as shown in \cref{fig:fig2}, which demonstrates the accuracy of our model's responses to different types of user prompts.

\textbf{Few-Shot Result.} To further validate the effectiveness of FocusChat, we conduct few-shot experiments based on a zero-shot model on open-domain visual question-answering datasets ActivityNet-QA, MSVD-QA, MSRVTT-QA, and the multiple-choice dataset Next-QA. As shown in \cref{tab:few-shot}, we compare recent few-shot models on these datasets, many of which use quite large pre-training datasets. For instance, VideoCoCa achieves state-of-the-art performance on ActivityNet-QA with a pre-training dataset size as large as 4.8B, whereas our model only uses 0.72M pre-training data, still yields comparable results with SOTA performance. FocusChat gets 63.7 with GPT-3.5-turbo-0613 evaluation on MSVD-QA, 54.6 without GPT-3.5-turbo-0613 evaluation. The accuracy on Next-QA  is 68.20, second only to the SeViLA model. This demonstrates the significant potential of FocusChat.

\subsection{Ablation Study }
To verify the effectiveness and rationality of FocusChat's design, we conduct few-shot and zero-shot ablation experiments based on the ActivityNet-QA dataset.

\textbf{For few-shot ablation}, each experiment in \cref{tab:ablution} involves pre-training and fine-tuning, with the same data used as in the previous zero-shot experiments. Since our model architecture is similar to Video-LLaMA, we use it as a reference for comparison. The ``baseline" in \cref{tab:ablution} refers to replacing the trainable positional encoding in Video-LLaMA's video Q-Former with our proposed absolute positional encoding \textbf{VPE} and adding an LN layer to each layer of the video Q-Former, as shown in \cref{eq:cc}. This improvement increased the model accuracy from 45.07 to 47.34. Rows from 3 to 6 in  \cref{tab:ablution} show the impact of different alpha and beta parameter values in the STFM module on model performance. The optimal parameters were $\alpha$ = 1 and $\beta$ = 0, resulting in an accuracy of 48.93, an improvement of 1.59 over the baseline. This demonstrates that the effectiveness of the spatial-level feature filtering process aligning with prompt semantics. Rows of 7 and 8 in \cref{tab:ablution} introduce ablation of prompt-based temporal filtering~(PBTF) module. With this addition, FocusChat's accuracy increased to 49.4, validating the effectiveness of time-level feature filtering. Finally, to further confirm the effectiveness of \textbf{VPE}, we replaced the positional encoding in FocusChat with the original trainable positional encoding from Video-LLaMA. The accuracy dropped to 49.29, reaffirming the validity of the proposed positional encoding block.

\textbf{The zero-shot ablation experiment} was conducted to verify the impact to the number of vision tokens on model performance. This experiment was also performed on ActivityNet-QA. We train a FocusChat model with 16 vision tokens, denoted as FocusChat~(16). As shown in \cref{tab:num_token}, even with 16 vision tokens, 0.72M pretraining data, and 0.8M fine-tuning data, the model accuracy surpassed Video-LLaMA by 15.3, with only 5.5 points drop compared to FocusChat. This demonstrates that the effectiveness of our model in extracting visual features lowers the essential acount of vision tokens needed.
\section{Conclusion}
In this paper, we introduce FocusChat, a text-guided model that employs spatiotemporal information filtering to realize the efficiency of visual information. As far as we know, we are the first to explore redundant spatiotemporal information filtering of visual features by explicitly applying it to attention. The extracted visual tokens properly aligning with the user's input elevate the model's vision understanding capacity. Experiments confirm the effectiveness of FocusChat. In few-shot experiments, it obtained competitive performance on par with SOTA models, with only 10 percent of the pre-training data. It outperforms Video-LLaMA in zero-shot tasks with only 16 visual tokens used and a magnitude less data. Besides the research value, the proposed method will show its advantages in practical and constrained scenarios.

{\small
\bibliographystyle{ieee_fullname}
\bibliography{egbib}
}

\end{document}